\documentclass{article}
\usepackage{arxiv}
\usepackage{verbatim}  
\usepackage[normalem]{ulem}
\usepackage{pdfpages}
\usepackage{lmodern}
\usepackage[english]{babel}
\usepackage{graphicx}
\usepackage{float}
\usepackage{bbm}
\usepackage{amsfonts}
\usepackage{tipa}
\usepackage[scientific-notation=true]{siunitx}
\usepackage{pdflscape}
\usepackage{afterpage}
\usepackage{xcolor}
\usepackage{tabularx}
\usepackage{capt-of}
\usepackage[toc,page]{appendix}
\usepackage[font=small,format=plain,labelfont=bf,textfont=normal,justification=justified,singlelinecheck=false]{caption}
\usepackage{color}
\usepackage[utf8]{inputenc} 
\usepackage[T1]{fontenc}    
\usepackage{hyperref}       
\usepackage{url}            
\usepackage{booktabs}       
\usepackage{amsfonts}       
\usepackage{nicefrac}       
\usepackage{microtype}      
\usepackage{lipsum}

\begin{document}
\title{Metric learning for phoneme perception}

\author{
  Yair Lakretz \\
  Cognitive Neuroimaging Unit \\ NeuroSpin Center \\ Gif-sur-Yvette, France\\
  \texttt{yair.lakretz@cea.fr} \\
  \And
  Gal Chechik \\
  The Gonda Brain Research Center\\
  Bar Ilan University \\
  Ramat-Gan, Israel \\
  \texttt{gal.chechik@biu.ac.il} \\
  \AND
  Evan-Gary Cohen \\
  Department of Linguistics \\ Tel-Aviv University \\ 
  Tel-Aviv, Israel \\
  \texttt{evan@post.tau.ac.il} \\
  \And
  Alessandro Treves \\
  Cognitive Neuroscience \\ SISSA \\ 
  Trieste, Italy \\
  \texttt{ale@sissa.it} \\
  \And
  Naama Friedmann \\
  Language and Brain Lab \\ School of Education and Sagol School of Neuroscience \\ Tel-Aviv University\\
  Tel-Aviv, Israel \\
  \texttt{naamafr@post.tau.ac.il} \\
}

\maketitle

\begin{abstract}

Metric functions for phoneme perception capture the similarity structure among phonemes in a given language and therefore play a central role in phonology and psycho-linguistics. Various phenomena depend on phoneme similarity, such as spoken word recognition or serial recall from verbal working memory. This study presents a new framework for learning a metric function for perceptual distances among pairs of phonemes. Previous studies have proposed various metric functions, from simple measures counting the number of phonetic dimensions that two phonemes share (place-, manner-of-articulation and voicing), to more sophisticated ones such as deriving perceptual distances based on the number of natural classes that both phonemes belong to. However, previous studies have manually constructed the metric function, which may lead to unsatisfactory account of the empirical data. This study presents a framework to derive the metric function from behavioral data on phoneme perception using learning algorithms. We first show that this approach outperforms previous metrics suggested in the literature in predicting perceptual distances among phoneme pairs. We then study several metric functions derived by the learning algorithms and show how perceptual saliencies of phonological features can be derived from them. For English, we show that the derived perceptual saliencies are in accordance with a previously described order among phonological features and show how the framework extends the results to more features. Finally, we explore how the metric function and perceptual saliencies of phonological features may vary across languages. To this end, we compare results based on two English datasets and a new dataset that we have collected for Hebrew.
\end{abstract}
\section{Introduction}
A common processing malfunction in speech perception is the misperception of one phoneme as another in normal or noisy conditions, usually in the context of spoken words or sentences. For example, when mishearing someone saying ‘take it’ as ‘make it’. Notably, some phoneme substitutions are more likely than others. For example, putting aside non-phonolgical effects, it is more likely to mishear ‘take it’ as ‘fake it’ than as ‘make it’. One contributing factor for such confusions is that the pair of phonemes /t/-/f/ is more confusable with each other than the pair /t/-/m/. This raises the question: what makes one pair of phonemes more perceptually confusable, or similar \citep{Tversky1977, Shepard1987}, than another pair?

To addresses this question, various phoneme-similarity theories were proposed in the literature \citep[e.g.,][]{Tversky1977, Frisch1997}. Phoneme-similarity theories suggest a similarity function that calculates phoneme-similarity for a given pair of phonemes. Typically, the function is defined over phonological features, but it could also be defined over other properties, such as membership in natural classes \citep{Pierrehumbert1993}, or the number of phonological dimensions (voicing, place- and manner-of-articulation) on which two phonemes agree \citep[e.g., ][]{Bailey2005, mousikou2015masked}. A phoneme-similarity theory suggests an answer to the above question by explaining similarity in terms of a phonological theory. For example, the relatively high similarity between the phonemes /n/ and /d/ may be explained by the fact that these two phonemes share both place-of-articulation and voicing dimensions. However, a common limitation of phoneme-similarity theories is that the exact form of the similarity function is determined by the theoretician, to best account for phoneme-similarity data. Since phoneme-similarity theories were shown to have limitations in explaining empirical data \citep{Bailey2005}, having a more systematic way to determine the exact form of the similarity function may benefit phoneme-similarity theories.

This study explores such an approach, which suggests a framework to systematically deriving phoneme-similarity functions from empirical data. It differs from previous studies in that the similarity function is derived from the data by using learning algorithms, instead of relying on human ability to manually capture the data. According to this approach, only the general form of the similarity function is predefined, and its exact form is determined from the data, as explained below.

Importantly, the suggested framework allows exploration of additional questions. First, it allows to learn a phoneme-similarity function separately for each language. Phoneme-similarity in different languages may be better explained with different similarity functions. As we later show, this also allows to test whether a given phonological feature has the same perceptual saliency across languages. Second, the suggested framework allows to learn a phoneme-similarity function defined over different phonological theories. As we later show, different phonological-features theories can be explored under the suggested framework. Finally, additional questions regarding phoneme similarity, such as the role of asymmetry in phoneme perception and differences in phoneme-similarity between perception and production, can be explored under the suggested framework, as we later discuss.

In particular, this study addresses the following questions: first, we ask whether data-driven phoneme-similarity function can better predict unseen data compared to existing theories for phoneme similarity. This is a necessary result for data-driven phoneme-similarity functions to be of any interest. Second, we ask whether data-driven similarity functions can reveal perceptual differences among different phonological features. Finally, we ask whether the similarity function significantly differs across languages. To address this question, we collect phoneme-confusion data in Hebrew and conduct a comparison based on existing phoneme-confusion datasets in English. 

The rest of this section is organized as follows. Section 1.1 provides a brief history of the main developments of phonological theories, described according to two lines of research - auditory and gestural theories. Section 1.2 discusses phoneme-similarity theories and the empirical approach to phoneme similarity. Finally, section 1.3 presents in more detail the new approach this study suggests to the problem.

\subsection{Feature theories}
Speech sounds often exhibit the same behavior, grouping together into sound patterns in spoken languages. For example, in German, the singular and plural forms consistently differ by specific pairs of sounds, such as /b/-/p/ in the German word for 'robbery': Raube vs. Raub (final b pronounced as /p/); or for 'bath': /d/-/t/ in  Bäder  vs. Bad (final d pronounced as /t/), and /g/-/k/, /z/-/s/ and /v/-/f/ in other cases. All these different pairs of phonemes have a single property in common: in the plural form, the sounds (/bgdvz/) always include voicing - a vibration of the vocal chords during pronunciation, which has identifiable acoustic markers - whereas in the singular form, the corresponding sounds are unvoiced. Such groups of sounds are termed \textit{natural classes} and they are denoted by subphonemic features, such as [-voice] and [+consonantal]. Natural classes are typically affected, and affect other sounds, in the same way in the same environment, as in the above example. It was therefore suggested that phonological alterations, such as the devoicing process, act on subphonemic features rather than on the phonemic segments themselves. Interestingly, different languages may present similar to identical processes. For example, in Turkish, similarly to German, we find final devoicing in, e.g., 'kitabim' ('my book') vs. 'kitap' ('book'), or 'a\u{g}aim' ('my tree') vs. 'a\u{g}a\c{c}' ('tree'). Such similarities have led to the hypothesis that phonological features are universal and innate (\citealp{ChomskyHalle1968}, but see \citealp{mielke2008emergence}).

\paragraph{Auditory theories}
Around the mid of the 20$^{th}$ century, \citet{jakobson1951preliminaries} suggested that phonological patterns in all languages can be described with only a limited number of contrasting features, such as, voiced-unvoiced, strident-mellow, or nasal-oral. According to this theory, all features can be assigned a unique acoustic correlate, which in turn, correspond to specific articulatory parameters. The articulatory stage, however, is viewed as a mere mechanism to generate acoustic contrasts - "we speak to be heard in order to be understood" (ibid.), and features do not necessarily have a unique articulatory correlate. This line of research, emphasizing the acoustic aspect of distinctive features, was later developed by \textit{the quantal theory of speech perception} \citep{stevens1972quantal,stevens1989quantal}. This theory added to the field by characterizing the interactions between articulatory parameters of speech and their acoustic outcomes. It suggested that the mapping between articulation to acoustics is highly non-linear, with stable regions (small changes in the articulator have small acoustic effects) and unstable regions (small changes in the articulator generate large acoustic effects). The discontinuities around the unstable regions are used to define the boundaries between features. In this way, the theory accounts for a fundamental question in the field - why languages distinctly favor certain articulatory and acoustic dimensions in constructing their phoneme inventories, while avoiding others. The quantal theory explains this by defining optimality of features - a feature is optimal if it maximizes the difference between phonemes in the auditory space while preserving minimal articulatory effort. This phonological economy explains why features are universal, as they are dependent on biological properties of the human speech-production system, which are essentially the same for all humans.

\paragraph{Gestural theories}
From around the middle of the 20$^{th}$ century, a competing view to auditory theories was developed, called {\it The Motor Theory of Speech Perception} \citep{cooper1952some, liberman1967perception}. Its primary motivation stemmed from the observation that acoustic waveforms of synthetic speech have to be modified in order to produce a constant perception of the same phoneme in different phonological contexts. It was therefore concluded that the objects of speech perception cannot be found in acoustics, in contrast to the view of auditory theories, but should rather be anchored in articulatory gestures. Accordingly, during speech perception, the hearer extracts the intended phonetic gestures of the speaker \citep{liberman1985motor}, rather than acoustic cues. Phonemes are thus viewed as groups of gestural features. In a later theory in the same thread of research, giving emphasis to articulation as well, \citet{ChomskyHalle1968} suggested that phonological patterns across languages can be accounted for with an innate set of articulatory features. Importantly, they argue that what is perceived depends not only on the physical constitution of the signal but also on the linguistic knowledge of the hearer, introducing top-down and expectation processes into the field of speech perception. A further development of this line of research, called feature geometry \citep{clements1985, sagey1986representation}, suggested that speech sounds have an internal featural structure, rather than being comprised of a "bundle" of features. In its simple form, the theory describes the internal structure of segments in terms of a tree, whose nodes are features, higher levels are feature classes and, finally, the root node represents the entire object. Describing sounds with trees enabled the theory to account for phonological patterns in a more parsimonious way compared to previous theories. More recently, \citet{browman1990tiers, browman1992articulatory} suggested a model of gestural coordination known as \textit{Articulatory phonology}. Their underlying hypothesis is that spoken language is best described by patterns of coordinated articulatory gestures. The theory proposes to model phonetic and phonological patterns in terms of "articulatory scores", which are formal representations consisting of abstract gestures and their patterns of coordination.

\subsection{Theories for phoneme similarity}
\paragraph{Feature-based approach}
Phoneme similarity is traditionally explained by decomposing phonemes into subphonemic features, assuming the more features two phonemes share the more similar they are \citep{Tversky1977, Shepard1987, Cohen2009, Cohen2010}. Given a feature theory, a metric (distance) or a phoneme-similarity functions can be defined over pairs of phonemes. For example, given the three contrastive articulatory dimensions - voicing, place- and manner-of-articulation - a common similarity function in psycholinguistics is to count the number of dimensions on which two phonemes agree (e.g., \citealp{Bailey2005, mcmillan2010cascading, mousikou2015masked}). For instance, the phoneme /b/ is labial-plosive-voiced, and the phoneme /g/ is velar-plosive-voiced. The similarity score for these two phonemes is thus two. Other, more sophisticated, similarity measures have been proposed in the literature. These include a similarity measure based on the number of shared and non-shared features \citep{Pierrehumbert1993}, or on the number of shared and non-shared natural classes \citep{Frisch1997}. According to this last measure, similarity between two phonemes is the proportion of natural classes that are shared by the two phonemes, compared to the total shared and not-shared natural classes. Features are thus assigned weights according to their redundancy. Therefore, different features contribute differently in explaining perceptual distances, since redundant features affect similarity less than non-redundant features.

\paragraph{Experimental approach}
Another approach to phoneme similarity addresses the problem by quantifying perceptual phoneme similarities directly in an experiment. Typically, such experiments present noise-corrupted phonemes to subjects, collecting their perceptual judgments of phoneme identity \citep{NicelyMiller1955}. Confusion matrices are then derived from the data, mapping the likelihood that any given phoneme is confused with another phoneme. This approach has been widely used, however, there are several associated concerns that should be kept in mind. First, phoneme confusions may depend on the noise characteristics added to the signal. For example, results from phoneme-confusion experiments using white noise (e.g., \citealp{NicelyMiller1955, Luce1987}) may differ from those that use pink noise (e.g., \citealp{redford1999relative}), or speech-shaped ('babble') noise \citep{cutler2004patterns}. Another source of variability is due to context effects. These may have large effects on perceptual confusability, in particular, consonant confusions may be different when the consonant is in initial or final position of the word \citep{Luce1987, hura1992role, redford1999relative}. Finally, phoneme perception shows asymmetric effect \citep{kuhl1991human}, which was addressed by several feature theories. For example, the featurally underspecified lexicon (FUL) theory \citep{lahiri2002underspecified, lahiri2010distinctive} accounts for asymmetry effects by stating that some features are universally underspecified. For instance, the feature [coronal] is underspecified in the lexicon, which prevents mismatches between labial or dorsal sounds that are detected in the signal, and allows them to access the lexicon. In contrast, labial and dorsal features are specified, and so a coronal sound detected in the signal will mismatch both of these features, and cannot access the lexicon. In speech production, \citet{Frisch1997} claims that similarity neighborhoods in the mental lexicon affect asymmetries in sound substitutions.

\subsection{A metric-learning approach}
We now describe a general framework that we suggest as a new methodological approach to the problem of phoneme similarity. The approach allows the integration of various effects, such as context or asymmetry, into its framework, and is compatible with various feature theories - e.g., auditory or gestural, and with any choice of noise characteristics in phoneme-confusion experiments - white, pink or speech-shaped.

Traditionally, feature-based theories, such as the ones described above, craft a similarity or a metric function based on theoretical insights, however, current theoretical metrics do not fully account for empirical perceptual distances (e.g. \citealp{Bailey2005}). One reason for this may be that theoretical metrics misevaluate or underestimate the difference in contribution of each subphonemic feature to perceptual distances. 

In this study, we explore an approach that learns a metric function directly from the data, assuming only its general form. For this, we parameterize a general metric function and derive its values using learning algorithms. The metric function is defined over phonological features, based on a theory for phonological features, and its parameters are estimated to best explain empirical data. The approach can be therefore seen as bringing together the two approaches to phoneme similarity described above: the feature-based and the experimental approach to phoneme similarity. 

A geometric interpretation of this approach is illustrated in figure 1. Given a feature theory, phonemes can be represented as points in feature space, in which dimensions correspond to subphonemic features (figure 1 left). The set of all Euclidean distances between phonemes in this space likely differ from the set of the empirical ones (figure 1 right) from , e.g., a phoneme-confusion data. Given this discrepancy between the two structures, we look for an optimal mapping between the two spaces, such that the distances from the feature space will match as closely as possible the set of distances from the experiment (figure 1 middle arrow). This optimal mapping defines a metric function over pairs of phonemes. Importantly, once the optimal mapping is derived from the data, one can learn about the perceptually saliency of the various features. As an example, if the mapping is linear and represented by a (positive) diagonal matrix, the mapping amounts to stretchings or contractions of the feature dimensions. Dimensions that are relatively highly stretched would thus correspond to features that are more perceptually discriminative.

\begin{figure}
\includegraphics[width=15cm]{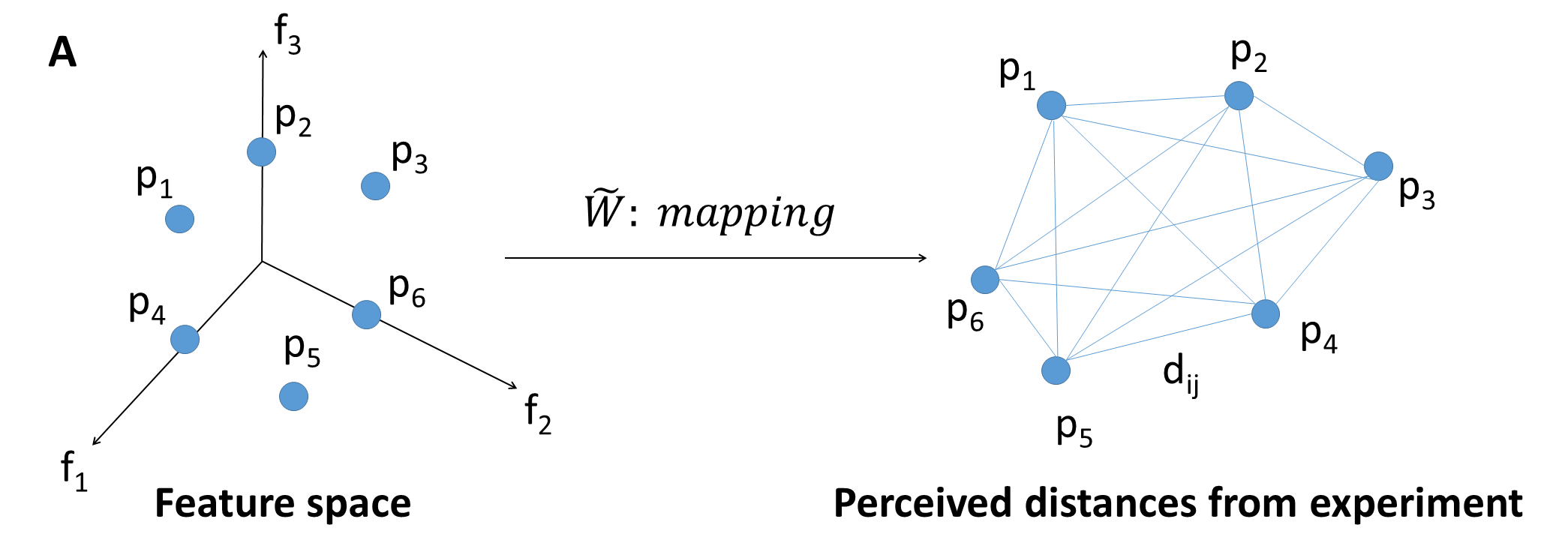}
\caption{An illustration of metric learning for phonemes: phonemes $p_i's$ are represented in a feature space (left), where each dimension corresponds to a feature. Perceptual distances as collected in an experiment $d_{ij}$ (right) are used to learn an optimal mapping (middle arrow), defined over the feature space. The mapping is optimal in the sense that it maps the phonemes from the feature space such that the distances between the mapped phonemes are as close as possible to the empirical ones.}
\end{figure}

We test our approach on two English phoneme-confusion datasets \citep{NicelyMiller1955, Luce1987} and a new dataset that we have collected for Hebrew. In all cases, we focus on consonant confusions at initial position and white noise in all experiments. In addition, we present and compare several alternatives to learn the metric function. For computational convenience, we assume a linear mapping and a symmetric metric function. However, the approach can be pursued with a removal of such assumptions, as we later discuss.
\section{Methods}
We describe learning a metric given a feature set and a phoneme-similarity dataset. Section 2.1 describes phoneme representations in feature space. Section 2.2 describes different methods for learning optimal mappings. Finally, Section 2.3 describes phoneme-confusion datasets that are used in this study, and the way perceptual distances were estimated.

\subsection{Representing phonemes in feature space}
We study phoneme representations in the feature space based on: (a) Articulatory features, and (b) Phonological features. See table 1 for the full list of phonemes.

\paragraph{Articulatory features} The first representation scheme uses 14 binary features, based on articulatory features: six place features - labial, dental, alveolar, palatal, velar and glottal, ordered according to the place along the vocal tract; Seven manner features - plosive, fricative, affricate, lateral, rhotic, nasal, glide, ordered by sonority; And a single voicing feature. As an example, the labial-plosive-unvoiced /p/ is represented  as (1, 0, 0, 0, 0, 0, 0, 1, 0, 0, 0, 0, 0, 0), where the first six components are for place of articulation, the next seven components are for manner and the last vector component is for voicing. Another example, the alveolar-nasal-voiced /n/ is represented as (0, 0, 0, 0, 1, 0, 0, 0, 0, 0, 0, 1, 0, 1).

\paragraph{Phonological features} Phonemes are represented along 12 binary phonological features: consonantal, continuant, strident, voicing, nasality, dorsal, anterior, labial, coronal, distributed and lateral. Continuing with the above examples, /p/ is represented here as (1, 0, 0, 0, 0, 0, 0, 0, 1, 0, 0, 0); and /n/ is represented as (1, 0, 0, 0, 1, 1, 0, 1, 0, 1, 0, 0). \mbox{} \\

Both representation schemes include the minimal set of features that is required to distinguish among all phonemes in the English and Hebrew datasets. Adding more features introduces redundancy among the features. For example, to distinguish between the sets of phonemes in the datasets, the sonority feature is not required in addition to the above list of features. Adding it would introduce redundancy with, e.g., the voicing feature, which would reduce the predictive power of the metric function and its interpretability.

\begin{landscape}

\begin{table}[H]
\centering 
\tiny
\begin{tabular}{|c|c|c|c|c|c|c|c|c|c|c|c|c||c|c|c|c|c|c|c|c|c|c|c|c|c|c|}
\hline

&\multicolumn{12}{|c|}{Phonological features}&\multicolumn{14}{|c|}{Articulatory features}\\
\hline
&	cn	&	ct	&	DR	&	st	&	vc	&	ns	&	dr	&	an	&	lb	&	cr	&	ds	&	lt	&	lb	&	dn	&	al	&	pa	&	vl	&	gl	&	pl	&	af	&	fr	&	ns	&	lt	&	rt	&	gd	&	vc	\\
\hline

${p}^{\star\dagger\ddagger}$	&	+	&	-	&	-	&	-	&	-	&	-	&	-	&	-	&	+	&	-	&	-	&	-	&	+	&	-	&	-	&	-	&	-	&	-	&	+	&	-	&	-	&	-	&	-	&	-	&	-	&	-	\\
$b^{\star\dagger\ddagger}$	&	+	&	-	&	-	&	-	&	+	&	-	&	-	&	-	&	+	&	-	&	-	&	-	&	+	&	-	&	-	&	-	&	-	&	-	&	+	&	-	&	-	&	-	&	-	&	-	&	-	&	+	\\
$t^{\star\dagger\ddagger}$	&	+	&	-	&	-	&	-	&	-	&	-	&	-	&	+	&	-	&	+	&	-	&	-	&	-	&	-	&	+	&	-	&	-	&	-	&	+	&	-	&	-	&	-	&	-	&	-	&	-	&	-	\\
$d^{\star\dagger\ddagger}$	&	+	&	-	&	-	&	-	&	+	&	-	&	-	&	+	&	-	&	+	&	-	&	-	&	-	&	-	&	+	&	-	&	-	&	-	&	+	&	-	&	-	&	-	&	-	&	-	&	-	&	+	\\
$k^{\star\dagger\ddagger}$	&	+	&	-	&	-	&	-	&	-	&	-	&	+	&	-	&	-	&	-	&	-	&	-	&	-	&	-	&	-	&	-	&	-	&	-	&	+	&	-	&	-	&	-	&	-	&	-	&	-	&	-	\\
$g^{\star\dagger\ddagger}$	&	+	&	-	&	-	&	-	&	+	&	-	&	+	&	-	&	-	&	-	&	-	&	-	&	-	&	-	&	-	&	-	&	-	&	-	&	+	&	-	&	-	&	-	&	-	&	-	&	-	&	+	\\
\textipa{tS}$^{\star}$	&	+	&	-	&	+	&	+	&	-	&	-	&	-	&	-	&	-	&	+	&	+	&	-	&	-	&	-	&	-	&	+	&	-	&	-	&	-	&	+	&	-	&	-	&	-	&	-	&	-	&	-	\\
\textipa{dZ}$^{\star}$	&	+	&	-	&	+	&	+	&	+	&	-	&	-	&	-	&	-	&	+	&	+	&	-	&	-	&	-	&	-	&	+	&	-	&	-	&	-	&	+	&	-	&	-	&	-	&	-	&	-	&	+	\\
$f^{\star\dagger\ddagger}$	&	+	&	+	&	-	&	-	&	-	&	-	&	-	&	-	&	+	&	-	&	-	&	-	&	+	&	-	&	-	&	-	&	-	&	-	&	-	&	-	&	+	&	-	&	-	&	-	&	-	&	-	\\
$v^{\star\dagger\ddagger}$	&	+	&	+	&	-	&	-	&	+	&	-	&	-	&	-	&	+	&	-	&	-	&	-	&	+	&	-	&	-	&	-	&	-	&	-	&	-	&	-	&	+	&	-	&	-	&	-	&	-	&	+	\\
\textipa{T}$^{\star\dagger}$	&	+	&	+	&	-	&	-	&	-	&	-	&	-	&	+	&	-	&	+	&	-	&	-	&	-	&	+	&	-	&	-	&	-	&	-	&	-	&	-	&	+	&	-	&	-	&	-	&	-	&	-	\\
\textipa{D}$^{\star\dagger}$	&	+	&	+	&	-	&	-	&	+	&	-	&	-	&	+	&	-	&	+	&	-	&	-	&	-	&	+	&	-	&	-	&	-	&	-	&	-	&	-	&	+	&	-	&	-	&	-	&	-	&	+	\\
$s^{\star\dagger\ddagger}$	&	+	&	+	&	-	&	+	&	-	&	-	&	-	&	+	&	-	&	+	&	-	&	-	&	-	&	-	&	+	&	-	&	-	&	-	&	-	&	-	&	+	&	-	&	-	&	-	&	-	&	-	\\
$z^{\star\dagger\ddagger}$	&	+	&	+	&	-	&	+	&	+	&	-	&	-	&	+	&	-	&	+	&	-	&	-	&	-	&	-	&	+	&	-	&	-	&	-	&	-	&	-	&	+	&	-	&	-	&	-	&	-	&	+	\\
\textipa{S}$^{\star\dagger\ddagger}$	&	+	&	+	&	-	&	+	&	-	&	-	&	-	&	-	&	-	&	+	&	+	&	-	&	-	&	-	&	-	&	+	&	-	&	-	&	-	&	-	&	+	&	-	&	-	&	-	&	-	&	-	\\
\textipa{Z}$^{\dagger}$	&	+	&	+	&	-	&	+	&	+	&	-	&	-	&	-	&	-	&	+	&	+	&	-	&	-	&	-	&	-	&	+	&	-	&	-	&	-	&	-	&	+	&	-	&	-	&	-	&	-	&	+	\\
$h^{\star\dagger\ddagger}$	&	+	&	+	&	-	&	-	&	-	&	-	&	-	&	-	&	-	&	-	&	-	&	-	&	-	&	-	&	-	&	-	&	-	&	+	&	-	&	-	&	+	&	-	&	-	&	-	&	-	&	-	\\
$m^{\star\dagger\ddagger}$	&	+	&	-	&	-	&	-	&	+	&	+	&	-	&	-	&	+	&	-	&	-	&	-	&	+	&	-	&	-	&	-	&	-	&	-	&	-	&	-	&	-	&	+	&	-	&	-	&	-	&	+	\\
$n^{\star\dagger\ddagger}$	&	+	&	-	&	-	&	-	&	+	&	+	&	-	&	+	&	-	&	+	&	-	&	-	&	-	&	-	&	+	&	-	&	-	&	-	&	-	&	-	&	-	&	+	&	-	&	-	&	-	&	+	\\
\textipa{N}$^{\star}$	&	+	&	-	&	-	&	-	&	+	&	+	&	+	&	-	&	-	&	-	&	-	&	-	&	-	&	-	&	-	&	-	&	+	&	-	&	-	&	-	&	-	&	+	&	-	&	-	&	-	&	+	\\
$l^{\star\ddagger}$	&	+	&	+	&	-	&	-	&	+	&	-	&	-	&	+	&	-	&	+	&	-	&	+	&	-	&	-	&	+	&	-	&	-	&	-	&	-	&	-	&	-	&	-	&	+	&	-	&	-	&	+	\\
\textipa{R}$^{\star}$	&	+	&	+	&	-	&	-	&	+	&	-	&	-	&	-	&	-	&	+	&	-	&	-	&	-	&	-	&	+	&	-	&	-	&	-	&	-	&	-	&	-	&	-	&	-	&	+	&	-	&	+	\\
$w^{\star}$	&	-	&	+	&	-	&	-	&	+	&	-	&	+	&	-	&	+	&	-	&	-	&	-	&	+	&	-	&	-	&	-	&	-	&	-	&	-	&	-	&	-	&	-	&	-	&	-	&	+	&	+	\\
$j^{\star\ddagger}$	&	-	&	+	&	-	&	-	&	+	&	-	&	-	&	-	&	-	&	+	&	+	&	-	&	-	&	-	&	-	&	+	&	-	&	-	&	-	&	-	&	-	&	-	&	-	&	-	&	+	&	+	\\
\textipa{X}$^{\ddagger}$	&	+	&	+	&	-	&	-	&	-	&	-	&	+	&	-	&	-	&	-	&	-	&	-	&	-	&	-	&	-	&	-	&	+	&	-	&	-	&	-	&	+	&	-	&	-	&	-	&	-	&	-	\\
\textipa{ts}$^{\ddagger}$	&	+	&	-	&	+	&	+	&	-	&	-	&	-	&	+	&	-	&	+	&	-	&	-	&	-	&	-	&	+	&	-	&	-	&	-	&	-	&	+	&	-	&	-	&	-	&	-	&	-	&	-	\\
\textipa{K}$^{\ddagger}$	&	+	&	+	&	-	&	-	&	+	&	-	&	+	&	-	&	-	&	-	&	-	&	-	&	-	&	-	&	-	&	-	&	+	&	-	&	-	&	-	&	-	&	-	&	-	&	+	&	-	&	+	\\

\hline

\end{tabular}
\caption{Phonological and articulatory features of all phonemes in the datasets. cn=consonantal; ct=continuant; DR=delayed release; st=strident; vc=voicing; ns=nasal; dr=dorsal; an=anterior; lb=labial; cr=coronal; ds=distributed; lt=lateral; dn=dental; al=alveolar; pa=postalveolar; vl=velar; gl=glottal; pl=plosive; af=affricate; fr=fricative; lt=lateral; rt=rhotic; gd=glide. $^{\star}$Luce, $^{\dagger}$N\&M, $^{\ddagger}$Hebrew dataset.}
\end{table}
\end{landscape}

\subsection{Metric learning}
We explore two approaches to learn a metric function: (1) viewing the problem as a least-squares problem and (2) viewing the problem as a large-margin ranking problem \citep{weinberger2006distance, Chechik2010}. We begin by providing a formal definition of the problem, describing the terms, variables and parameters of the proposed model.

Let $A$ be a set, a metric function $d: A \times A \to \mathbb{R}$ is a non-negative symmetric function over pairs of elements of $A$, which assigns a value (a distance) to every pair in the set, and satisfies the triangle inequality and that for every element $x$ in $A$: $d(x, x) = 0$.

In our case, the elements of the set are phonemes $A=\{\bar{p}^1, ..., \bar{p}^{n_p}\}$, where $n_p$ is the number of phonemes. We describe each phoneme as a vector of size $n_f$, $\bar{p}^i \in \mathbb{R} ^{n_f}$, and define the following metric $d_{ij}(\bar{p}^i, \bar{p}^j)$ over pairs of phonemes:
\begin{equation}
    d_{ij}(\bar{p}^i, \bar{p}^j) = (\bar{p}^i - \bar{p}^j)^T W (\bar{p}^i - \bar{p}^j),
\end{equation}

where $W \in \mathbb{R} ^{n_f \times n_f}$ is a positive semi-definite (PSD) real-valued matrix. This function is a proper metric since it satisfies all conditions: non-negativity (PSD), symmetry, the triangle inequality, and that the distance between a phoneme to itself is always zero. Learning such metrics from data can be hard in the general case when the dimensionality of the input samples is large, calling for efficient optimization \cite{davis-et-al-icml-2007,atzmon2015learning,liu2015similarity}.

Geometrically, this metric can be described as: (1) performing a linear mapping on a given pair of vectors from the feature space; and (2) calculating the square Euclidean distance between the results. To see this, note that a PSD matrix $W$ can be decomposed such that $W = \widetilde{W}^T\widetilde{W}$. Given a pair of phoneme vectors, $\bar{p}^i$ and $\bar{p}^i$, the mapped vectors are $\widetilde{W}\bar{p}^i$ and $\widetilde{W}\bar{p}^j$, and the square Euclidean distance between a pair of mapped phonemes is therefore: $||\widetilde{W}\bar{p}^i - \widetilde{W}\bar{p}^j||^2 = (\bar{p}^i - \bar{p}^j)^T\widetilde{W}^T\widetilde{W}(\bar{p}^i - \bar{p}^j)$, which is the right-hand side of Eq(1).

The free parameters of the model are the elements of $W$. Given a set of perceived distances ${D_{ij}}$ between phonemes measured in an experiment, an optimal set of model parameters $W^\ast$ would bring the model distances $d_{ij}$ as close as possible to the empirical distances $D_{ij}$. Formally, we define the optimization problem:
\begin{equation}
    W^\ast = \min_{W \in S^n_+}{{\sum_{{i} < {j}}{(d_{ij} - D_{ij})^2}}},
\end{equation}
where the sum is over all ${n_p \choose 2}$ pairs of phonemes, and $S^n_+$ denotes the set of $n \times n$ PSD matrices. We describe below two methods to solve the optimization problem. The first reduces the problem into a least-squares (LS) problem, the second reduces it into a large-margin ranking problem (OASIS, \citealp{Chechik2010}). We also test a variant of each of the methods, by adding a diagonal constraint on the weight matrix (section 2.2.3), thus having four methods in total. We refer to these four ways of learning a metric function: LS, diagonal-LS, OASIS, diagonal-OASIS.

\subsubsection{Method A: A least-squares formulation}
Formulating the problem as a least-squares problem will allow the use of standard tools to solve the problem. We now show how the problem of metric learning as defined in section 2.2, can be described as a least-squares problem. We define a loss function $\mathcal{L}$ that correspond to the optimization problem in Eq(2).
\begin{equation}
    \mathcal{L} = \sum_{{i} < {j}}{((\bar{p}^i - \bar{p}^j)^TW(\bar{p}^i - \bar{p}^j) - D_{ij})} + \lambda||w||_1^2,
\end{equation}
where L1-regularization is added to the loss function in its second term. Adding L1-regularization has two advantages. First, it mitigates overfitting to the empirical training data. Second, L1-regularization results in sparser solutions, i.e., a larger number of parameters have values equal to zero \citep{Tibshirani1996}. Importantly, in our case, where feature weights are derived from data collected from a perceptual-discrimination task, non-zero weights that endured this penalty are those that are necessary for explaining the measured perceptual distances. Features that have higher weight are therefore more perceptually discriminative.

Rewriting the first term in the objective function, we get a classic LS formulation in the following matrix notation: $||\widetilde{P}\bar{w} - D||_2^2 + \lambda||w||_1^2$, where $\bar{w}$ is an $n_f^2$-dimensional vector, and $\widetilde{P}$ is an ${n_p \choose 2} \times n_f^2$ matrix. A row of $\widetilde{P}$ corresponds to the distance between the $i^{th}$ and $j^{th}$ phonemes, and a column corresponds to the weight of the $k^{th}$ and $l^{th}$ features from the quadratic form in Eq(1). The elements of $\widetilde{P}$ are therefore of the form: $(p_k^i-p_k^j)(p_l^i-p_l^j)$.

To solve the L1-regularized least-squares problem, we use the LASSO algorithm \citep{Tibshirani1996}. The weight of the regularizer, $\lambda$, was determined on a held out set, by comparing predictions of models with 18 different values of $\lambda$ in the range $[10^{-3}, 10^5]$. We selected the value that maximized prediction accuracy on a validation set. Note that due to the inequality constraint the resulting weight matrix has positive values, which in the case of a diagonal matrix (section 2.2.3) is guaranteed to be PSD. We refer to this method of learning a metric function as {\bf LS} for "least square".

In addition, we test a variant of this method in which the parameter matrix $W$ is constrained to be diagonal (see section 2.2.3). To ensure that the parameter matrix is a PSD matrix we constrain all its values to be non-negative using the constrained LASSO algorithm \citep{koh2007interior}. In the case of general (non-diagonal) matrices,  we project the resulting matrix onto the sub-space of PSD matrices using spectral decomposition, namely, by setting to zero the weights of eigenvectors that have negative eigenvalues.

\subsubsection{Method B: A large-margin ranking problem} The goal is to learn a metric function $d(p^i, p^j)$ as in Eq(1) that obeys the following condition \citep{Chechik2010}. Any given pair of phonemes ($p^i, p^i_+$) that are perceptually closer than another pair ($p^i, p^i_-$), should also by closer according to the learned metric function, formally, it is:
\begin{equation}
\forall{p^i, p^i_+, p^i_-}: D(p^i, p^i_+)<D(p^i, p^i_-) \rightarrow d(p^i, p^i_+)<d(p^i, p^i_-).
\end{equation}
The hinge-loss function to the constraint in Eq(4) together with a margin is:
\begin{equation}
    l_W(p^i, p^i_+, p^i_-) = max\{0, 1 + d(p^i, p^i_+) - d(p^i, p^i_-)\}.
\end{equation}
The optimization problem and the algorithm are then solved as in \citet{Chechik2010}, with a single projection onto the PSD matrix sub-space at the end of the training, using spectral decomposition. We refer to this method as OASIS.

Note that when using the LS method, the exact values from the empirical data are used to learn the parameters of the model $W$, whereas in OASIS, the data values are used for ranking the empirical distances between phonemes. It is an empirical matter which of the two methods better fits our problem, we therefore explore metric learning of phonemes with both LS and OASIS.

\subsubsection{Diagonal constraint} In addition, we test a version of the above two methods when a diagonal-constraint on the parameter matrix $W$ is added. That is, the off-diagonal weights in $W$ are set to zero. The advantage of adding a diagonal constraint is that the model has fewer parameters and it is thus less prone to overfitting. On the other hand, too simple a model may not be able to capture the full structure of the data. We test each method with and without a diagonal constraint. 

\subsubsection{Positive-semi definite } 

\subsection{Baseline methods}
We compare the predictions of the models to three baselines: \textit{uniform weights}, \textit{PMV} \citep{Bailey2005, ladefoged2014course} and \textit{Frisch similarity \citep{Frisch1997}}.

\paragraph{Baseline \#1: Uniform weights} $W$ is the identity matrix, with all weights set to 1 on the diagonal, and all other weights to zero.

\paragraph{Baseline \#2: Place-Manner-Voicing (PMV)} A common measure of phoneme similarity counts the number of articulatory dimensions - place, manner and voicing - that are shared by two phonemes. Formally,

\begin{equation}
S_{PMV}(p_i, p_j) = \delta_{place(p_i),place(p_j)} + \delta_{manner(p_i),manner(p_j)} + \delta_{voicing(p_i),voicing(p_j)},
\end{equation}

where $\delta_{i,j}$ is the Kronecker delta.

\paragraph{Baseline \#3: Frisch similarity \citep{Frisch1997}} Similarity is calculated based on shared and non-shared natural classes, instead of features:
 
\begin{equation}
\resizebox{.9\hsize}{!}{$S_{FRISCH}(p_i, p_j) = \frac{\#(shared \quad natural \quad classes)}{\#(shared \quad natural \quad classes) + \#(non-shared \quad natural \quad classes)}$}.
\end{equation}

By introducing natural classes, similarity becomes dependent on the redundancy of the features. Non-redundant features have more affect on similarity than redundant ones. In particular, a totally redundant feature adds no new natural classes and therefore does not affect similarity. This introduces a weighing to the similarity function that is based on redundancy. \mbox{} \\

\subsection{Perceptual distances from confusion data}
We now describe how the perceived distances ${D_{ij}}$ are estimated from empirical data. First, phoneme similarity is calculated from confusion data following Shepard's method:

\begin{equation}
    S_{ij}=\frac{p_{ij}+p_{ji}}{p_{ii}+p_{jj}},
\end{equation}

where $p_{ij}$ is the proportion of times that $i$ tokens were called $j$. Perceived distances are then calculated following Shepard's law \citep{Shepard1987, Johnson2004}:

\begin{equation}
    D(p^i, p^j) = -ln(S(p^i, p^j)).
\end{equation}

This assumes that the relationship between perceptual distances and similarity is exponential. \citet{Ennis1988} showed that this law can account for various empirical findings about object confusion (in same-different tasks) if decisions between objects are described in a probabilistic framework. That is, percepts are treated as probabilistic and Shepard's law is applied at the stage of decision between objects.

\subsection{Model evaluation}
To compare the quality of the proposed methods, we evaluate the predictive power of the metric function of each method using a cross-validation procedure. First, for each method, we learn a metric function from a subset of the data. We then assess the prediction of the learned function on left-out data. We repeat this by leaving out a different subset of the data each time, finally averaging the resulting predictions across all left-out subsets. More specifically, we use a leave-one-phoneme-out procedure, where each time, we leave out all perceived distances involving one phoneme. We then repeat this for all phonemes. Finally, we report the average of all resulting values. 

We use the Spearman rank correlation coefficient as a prediction measure. For each left-out phoneme $p$, we calculate the Spearman correlation $\rho_{d^p, D^p}$, between the model-predicted distances to the left-out phoneme $d^p$, and the empirical ones $D^p$. The final evaluation measure of model prediction is the average of this measure across all left-out phonemes $\bar{\rho} = \frac{1}{n_p}\sum_{p=1}^{n_p}\rho_{d^p, D^p}$. According to the distribution of the values $\rho_{d^p, D^p}$ we evaluate statistical significance between different methods.
\section{The data}
Confusion errors have traditionally been used as a measure for perceptual phoneme similarity, based on the idea that the more confusable two phonemes are the more similar they are. Three phoneme-confusion datasets are explored in this study: (1) the dataset from \citet{NicelyMiller1955}, (2) The dataset from \citet{Luce1987}, and (3) results from a new dataset collected from the experiment in Hebrew.

\subsection{The Nicely and Miller dataset} The classic work by \citet{NicelyMiller1955} analyzed auditory confusion between pairs of 16 consonant phonemes of English with added white noise at different signal-to-noise ratios (SNRs). The noise-corrupted phonemes were presented to subjects in a classification-task paradigm, in a consonant-vowel (CV) form, and the effect of low- and high-pass filtering on phoneme confusion was explored. The resulting confusion matrices from the experiments are provided in their paper. In what follows, we refer to this dataset as the \textit{N\&M dataset}, focusing on their -12dB SNR confusion matrix. Using tools from information theory, N\&M analysed the confusion matrices to see how subphonemic features - voicing, nasality, affrication, duration and place of articulation - affect phoneme confusion, describing subphonemic features as separated information-transfer channels. Their results suggest that the voicing and nasality features are less affected by random noise than the other three features tested: affrication, which distinguishes between /f \textipa{T} s \textipa{S} v \textipa{D} z \textipa{Z}/ and /p t k b d g m n/; duration, which distinguishes between /s \textipa{S} z \textipa{Z}/ and the other phonemes; and place of articulation with three-valued classification - front, middle and back. Affrication and duration were found to be more affected by random noise, but less than the place of articulation feature, which was the most affected by random noise. 

\subsection{The Luce dataset} The work carried by \citet{Luce1987} extended the work by N\&M in several aspects. First, the segment inventory was extended from 16 phonemes to 24 phonemes (see Table 1 for a full list of phonemes). Second, it used words instead of CV syllables, which is a more natural way of presenting stimuli. Words in the experiment only differed by the initial or the final consonant. Auditory confusion was analyzed for these two cases - initial and final - separately. Similarly to N\&M, auditory confusion was tested at different levels of SNR, using white noise. In what follows, we refer to this dataset as the \textit{Luce dataset}. To have higher confusion rate, we focused on initial-position confusions at the highest available noise level \citep{Redford1999}.

\subsection{A new Hebrew dataset}
To evaluate the perceptual confusability of Hebrew phonemes, we collected data from native Hebrew speakers.

\paragraph{Participants}
Thirty-two native, monolingual, Hebrew speaking participants (13 males, 19 females), ages: 21-35 (mean - 27.1), participated for course credit.

\paragraph{Stimuli}
All stimuli were recorded in an anechoic chamber with a RØDE NT2-A microphone and a Metric Halo MIO2882 audio interface, at a sampling rate of 44.1kHz. Stimuli were generated by two adult male native speakers of Hebrew. The total number of stimuli was 38 (19 phonemes * 2 speakers). Length and pitch (by semi-tone intervals) were compared across recorded tokens to choose the most highly comparable stimulus-types. This was done by looking at differences in timeline arrangement, using built-in pitch tracker in a commercial software (\textit{Logic Pro X}). Further cleaning of noise residues in high resolution mode was done using the \textit{Waves X-Noise} software. Next, all stimuli were added white noise at -12 SNR and normalized to the same RMS using a commercial software package (MATLAB, \textit{The MathWorks Inc., Natick, MA, 2000}).
\paragraph{Paradigm}
Stimuli were presented to participants in triplets, in an AXB manner, in which each token was a CV syllable. The vowel in all syllables was always /a/ since while front vowels (/i/ and /e/) may trigger fronting or palatalization, and non-low back vowels (/u/ and /o/) cause lip rounding, the low vowel /a/ has little effect on the adjacent consonantal phonemes \citep{ladefoged2014course}. At each trial, participants were asked to judge whether the middle phoneme (X) is the same as the first phoneme (A) or the last phoneme (B) by pressing one of two keys on the computer keyboard.

For each pair of phonemes {A, B}, all four possible triplets were presented to the participants: (1) AAB, (2) BAA, (3) ABB, and (4) BBA. If the participant was mistaken on either (1) or (2), the result was marked as a confusion of phoneme A with B. If the participant was mistaken on either (3) or (4), the result was marked as a confusion of the phoneme B with A. 

Sessions were conducted in a quiet room. The stimuli were presented in random order via earphones (Sennheiser HD 280 PRO). The Graphical User Interface (GUI) of the experiment was created with MATLAB, and ran on a laptop (Lenovo Yoga 2 Pro). A training session with 10 trials preceded the experiment. During the experiment, a pause was offered by the GUI to the participant every 20 trials. Participants could then decide to continue by pressing a key on the keyboard.

Outliers were discarded based on response time (RT) as follows. For each subject, the mean RT and standard deviation (SD) were calculated across all trials. Trials with RT above the mean plus three SDs were discarded.

Table 2 provides the resulting confusion matrix for all phonemes. The set of phonemes included 19 phonemes (see table 1 for the full list). In what follows, we refer to this dataset as the \textit{Hebrew dataset}.

\begin{landscape}
\renewcommand{\arraystretch}{0.8}
\begin{table}[H]
\centering
 \begin{tabular}{|c||c|c|c|c|c|c|c|c|c|c|c|c|c|c|c|c|c|c|c||c|}
\hline

Ph  & \textipa{b}	&	\textipa{g}	&	\textipa{d}	&	\textipa{h}	&	\textipa{v}	&	\textipa{z}	&	\textipa{X}	&	\textipa{t}	&	\textipa{j}	&	\textipa{k}	&	\textipa{l}	&	\textipa{m}	&	\textipa{n}	&	\textipa{s}	&	\textipa{f}	&	\textipa{p}	&	\textipa{ts}	&	\textipa{K}	&	\textipa{S} & Total \\
\hline
\hline
b	&	282	&	8	&	4	&	6	&	14	&	10	&	15	&	10	&	4	&	14	&	3	&	4	&	4	&	11	&	9	&	5	&	10	&	4	&	18	&	435 \\
g	&	8	&	279	&	5	&	2	&	8	&	6	&	6	&	12	&	4	&	5	&	6	&	2	&	4	&	5	&	6	&	4	&	0	&	5	&	18	&	385 \\
d	&	10	&	12	&	213	&	7	&	7	&	8	&	10	&	4	&	3	&	15	&	3	&	0	&	9	&	7	&	10	&	6	&	12	&	1	&	9	&	346 \\
h	&	8	&	2	&	3	&	385	&	5	&	4	&	9	&	7	&	1	&	11	&	0	&	19	&	9	&	1	&	7	&	9	&	7	&	15	&	8	&	510 \\
v	&	5	&	2	&	1	&	6	&	215	&	7	&	2	&	10	&	5	&	3	&	4	&	5	&	1	&	10	&	3	&	0	&	5	&	9	&	10	&	303 \\
z	&	7	&	8	&	11	&	4	&	6	&	250	&	3	&	8	&	3	&	4	&	5	&	8	&	5	&	12	&	2	&	0	&	5	&	9	&	3	&	353 \\
\textipa{X}	&	13	&	7	&	7	&	5	&	12	&	5	&	323	&	4	&	2	&	6	&	3	&	6	&	7	&	2	&	2	&	6	&	7	&	2	&	1	&	420 \\
t	&	2	&	6	&	17	&	0	&	2	&	4	&	3	&	310	&	2	&	0	&	6	&	0	&	0	&	6	&	1	&	6	&	10	&	5	&	6	&	386 \\
j	&	1	&	6	&	6	&	1	&	4	&	6	&	1	&	1	&	380	&	1	&	8	&	2	&	4	&	3	&	0	&	0	&	1	&	1	&	4	&	430 \\
k	&	7	&	12	&	0	&	4	&	28	&	9	&	3	&	15	&	8	&	292	&	4	&	1	&	8	&	16	&	10	&	5	&	13	&	4	&	13	&	452 \\
l	&	4	&	1	&	4	&	3	&	1	&	5	&	5	&	2	&	3	&	1	&	414	&	5	&	1	&	5	&	0	&	1	&	9	&	2	&	3	&	469 \\
m	&	3	&	2	&	5	&	10	&	4	&	0	&	7	&	1	&	2	&	3	&	1	&	307	&	2	&	3	&	6	&	0	&	1	&	3	&	2	&	362 \\
n	&	2	&	2	&	1	&	0	&	1	&	6	&	3	&	3	&	2	&	1	&	1	&	11	&	279	&	0	&	1	&	0	&	2	&	10	&	2	&	327 \\
s	&	8	&	25	&	10	&	4	&	5	&	3	&	6	&	7	&	7	&	6	&	3	&	4	&	6	&	301	&	8	&	6	&	20	&	6	&	17	&	452 \\
f	&	5	&	4	&	3	&	1	&	8	&	7	&	4	&	11	&	5	&	8	&	1	&	3	&	5	&	6	&	309	&	11	&	3	&	2	&	5	&	401 \\
p	&	7	&	4	&	1	&	5	&	10	&	3	&	12	&	2	&	2	&	7	&	1	&	14	&	10	&	5	&	5	&	370	&	2	&	2	&	1	&	463 \\
ts	&	8	&	30	&	8	&	2	&	11	&	1	&	5	&	9	&	10	&	1	&	2	&	2	&	6	&	0	&	9	&	7	&	343	&	5	&	10	&	469 \\
\textipa{K}	&	1	&	7	&	3	&	0	&	9	&	0	&	1	&	1	&	5	&	6	&	5	&	8	&	2	&	2	&	4	&	9	&	1	&	389	&	1	&	454 \\
\textipa{S}	&	9	&	9	&	23	&	5	&	7	&	10	&	1	&	7	&	5	&	7	&	7	&	4	&	8	&	8	&	4	&	5	&	11	&	6	&	278	&	414 \\
\hline
\hline
Total	&	390	&	426	&	325	&	450	&	357	&	344	&	419	&	424	&	453	&	391	&	477	&	405	&	370	&	403	&	396	&	450	&	462	&	480	&	409	&	7831 \\
 \hline
 \end{tabular}
\caption{Phoneme confusion in Hebrew between 19 phonemes (SNR -12, AXB paradigm). Each value in the matrix represents the confusion $C(i,j)$, which is the number of times that a phoneme in row $i$ was perceived as the phoneme in column $j$.}
\end{table}
\end{landscape}
\section{Results}
We describe the results of our experiments with the three datasets. Section 4.1 compares between four learning methods and several baselines. It identifies the best metric-learning method for further analysis. Based on this method, section 4.2 explores the perceptual saliencies of the phonological features in the English dataset. Finally, section 4.3 examines differences between the resulting metrics for Hebrew and English.

\subsection{Metric learning for phonemes}
We compare four methods for learning an optimal metric function: LS, diagonal-LS, OASIS, diagonal-OASIS, and compare them to three baseline methods - uniform weights, PMV and Frisch-similarity (sections 2.2, 2.3). To single out the best method we evaluate the predictive power of the metric function of each method.

\begin{figure}[H]
\vspace{.3in}
\makebox[\textwidth][c]{\includegraphics[width=\textwidth]{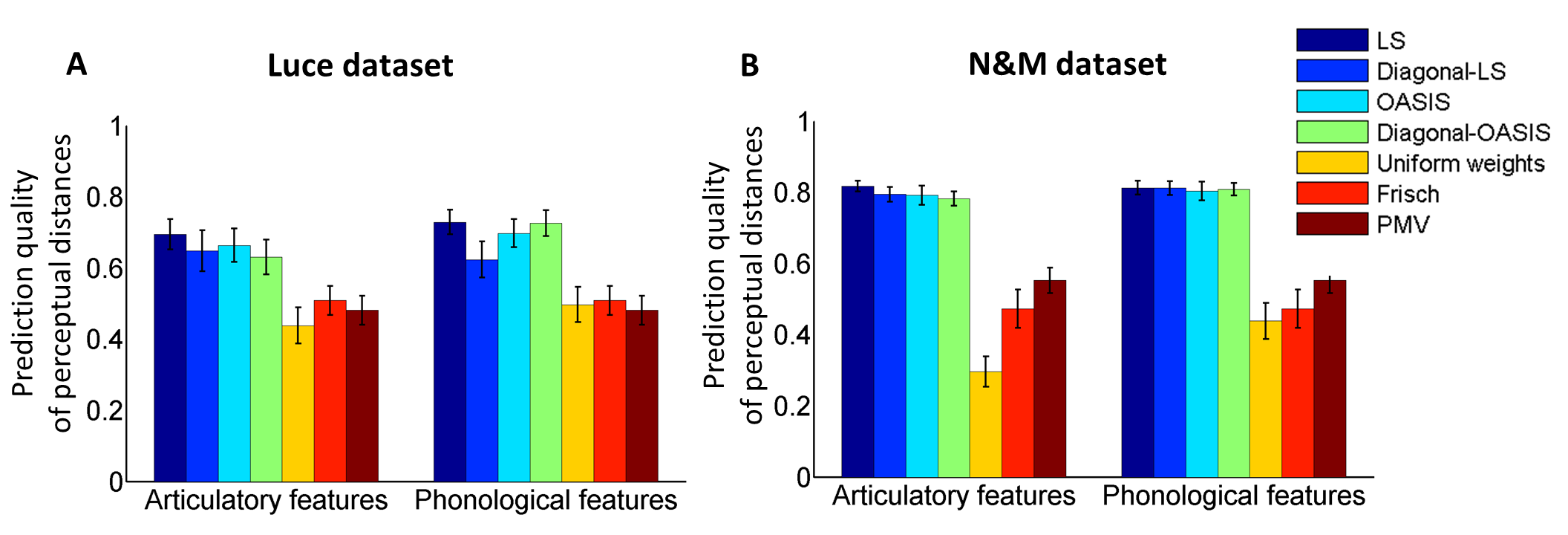}}
\caption{Prediction quality of perceptual distances. For each of the seven methods, the Spearman correlation between the predicted and empirical perceptual distances is calculated, averaged across left-out sets. Error bars represent SD across left-out sets.} 
\end{figure}

Figure 2 compares the average Spearman correlation between the seven methods. Results are shown for both confusion datasets: N\&M dataset, Luce dataset; and for the two feature theories discussed above: articulatory features and phonological features. Model prediction is evaluated in a cross-validation procedure (methods 2.5). The prediction of the model is defined as the average Spearman correlation across all validation sets, and the standard deviation is evaluated from the distributions across these sets. To determine the optimal regularizer size model predictions are evaluated on a validation set for various values of the regularization.

Results show that learning feature weights from data significantly improves prediction in comparison to baseline methods. For the N\&M dataset, all comparisons between the data-driven methods and the theoretical measures (Frisch, PMV, uniform weighing) are statistically significant ($p$-value$<10^{-4}$, two-tailed t-test - table 3). The four metric-learning methods demonstrate similar predictive power, as tested on these datasets. We therefore select a method for further analysis based on model parsimony and running time. Adding a diagonal constraint results in a more parsimonious model due to its smaller number of free parameters. We therefore choose to use diagonal-LS for the analyses in the rest of this study.

\begin{table}[]
    \centering
    \begin{tabular}{|c|c|c|c|c|}
        \hline
        \multicolumn{5}{|c|}{Luce - Articulatory features}\\
        \hline
         & LS & Diagonal-LS & OASIS & Diagonal-OASIS \\
         \hline
         Uniform weights & $<10^{-4}$ & $<10^{-4}$ & $<10^{-4}$ & $<10^{-4}$ \\
         Frisch similarity & 0.008 &    0.017 &    0.016 &    0.039 \\
         PMV & 0.003 &    0.034 &    0.006 &    0.031 \\
         \hline
         \multicolumn{5}{|c|}{Luce - Phonological features}\\
         \hline
         & LS & Diagonal-LS & OASIS & Diagonal-OASIS \\
         \hline
         Uniform weights & $<10^{-4}$ &  $<10^{-4}$ & $<10^{-4}$ & $<10^{-4}$ \\
         Frisch similarity & 0.033 &    0.012 &    0.020 &    0.011 \\
         PMV & 0.013 &    0.043 &    0.007 &    0.03 \\
         \hline
         
    \end{tabular}
    \caption{$p$-values computed using t-test for method comparison}
\end{table}

\subsection{Perceptually-discriminative features}
Given the sparsity constraint in the optimization problem (section 2.2.1), the weight of each subphonemic feature can be interpreted as its perceptual saliency. We thus examine the resulting weights from each dataset, to see what they reveal about the perceptually properties of the features.

Figure 3 shows the learned weights from the N\&M dataset for both the articulatory and phonological-features theories.

\begin{figure}
\makebox[\textwidth][c]{\includegraphics[width=\textwidth]{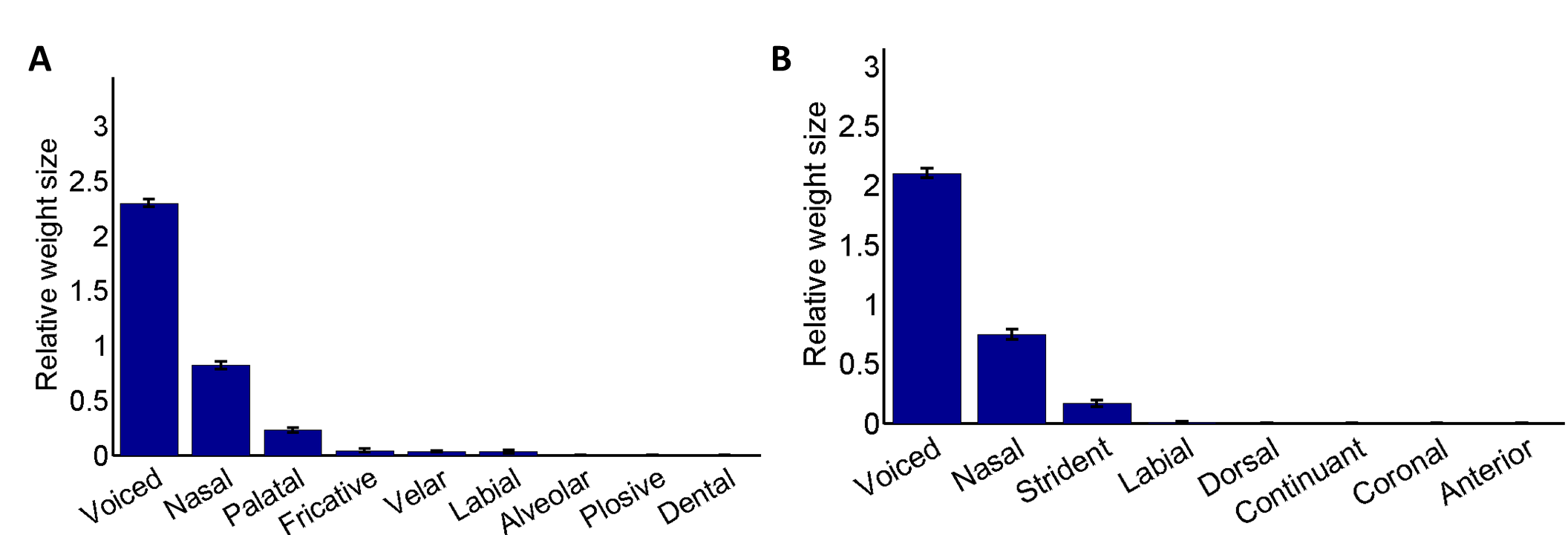}}
\caption{Weights of subphonemic features as derived from the Nicely and Miller dataset. (A) Articulatory features (B) Phonological features. Error bars represent SD across validation sets.}
\end{figure}

\paragraph{Articulatory features} Among the articulatory features, the voicing feature has the highest weight, followed by the nasality feature and then the palatal one. All differences are statistically significant ($p$-value$<10^{-4}$). Next in order are the fricative, velar and labial features, which are statistically comparable. Finally, the alveolar, plosive, and dental features have weights close to zero. Note that similarly to the conclusions from the N\&M study (section 3.1), the voicing and nasality features in English are less affected by random noise compared to place features, and other manner features.

\paragraph{Phonological features} When ordered by their weights, the phonological features are: voicing, nasality and the strident feature, in descending order (pairwise differences are statistically significant; t-test $p$-value$<0.05$). The other features, labial, dorsal, continuant, coronal and anterior have weights close to zero. 

Comparing the results of both feature theories, voicing has the highest weight in both cases, which is then consistently followed by nasality. In the third place are the palatal feature for the articulatory theory, and the strident feature for the phonological features theory. Examining this result, we note that the only palatal phonemes in the N\&M dataset are /\textipa{S}/ and /\textipa{Z}/, both are strident consonants. Similarly, when looking into the phonemes that share the strident feature, we find that /\textipa{S}/ and /\textipa{Z}/ are again the only strident phonemes in the N\&M dataset. It therefore seems that a common property of /\textipa{S}/ and /\textipa{Z}/ is the cause for the high weight of these features, and as suggested, it seems that /\textipa{S}/ and /\textipa{Z}/ being stridents, and in particular distributed stridents, is responsible for the result. We therefore summarize the results from the N\&M dataset about the perceptual discrimination of the leading features in the following order: voicing, nasality and distributed-stridents. 

We next examine what the Luce dataset reveals about perceptual-discriminative properties of subphonemic features. Figure 4 presents the learned weights from the Luce dataset for both features theories.

\begin{figure}[h]
\makebox[\textwidth][c]{\includegraphics[width=\textwidth]{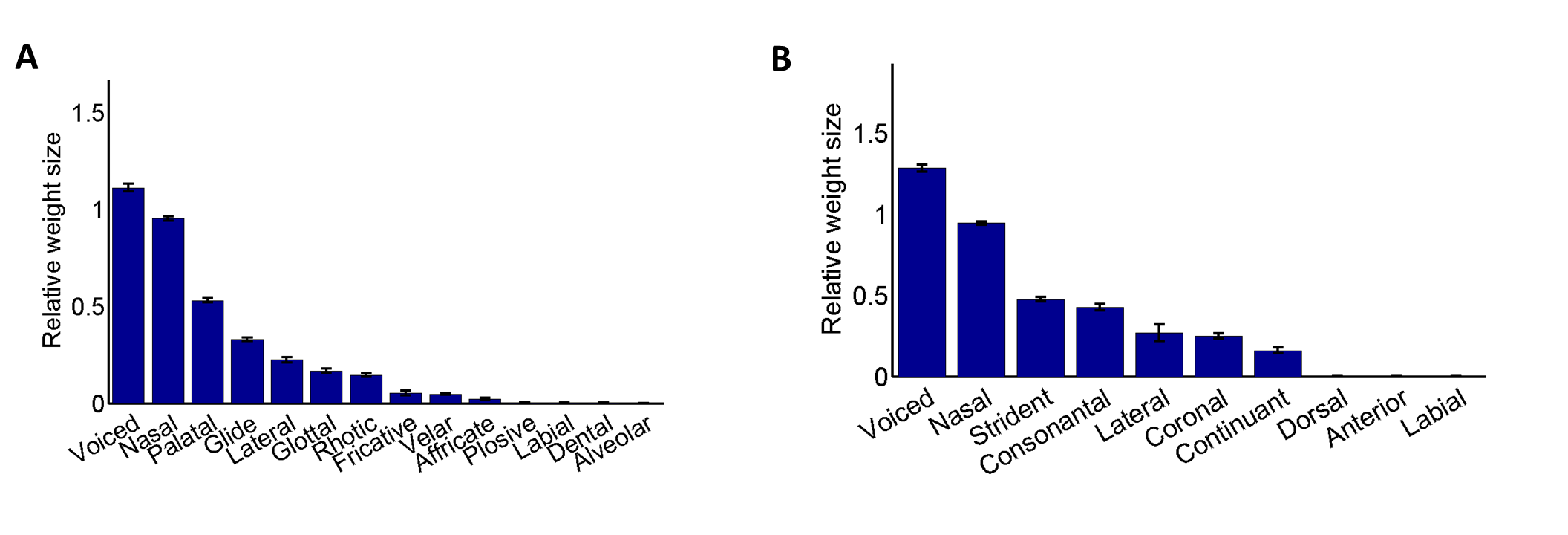}}
\caption{Weights of subphonemic features as derived from the Luce dataset. (A) Articulatory features (B) Phonological features (B). Error bars represent SD across validation sets.}
\end{figure}

\paragraph{Articulatory features} For the binary representation of articulatory features, the voicing feature has highest weight, followed by nasality, palatal, glide, lateral, glottal and rhotic features, in descending order. All differences are statistically significant (t-test $p$-value$<10^{-4}$) except for the glottal-lateral and glottal-rhotic differences. Next in order are the fricative and velar, which are statistically comparable. The rest of the features - affricate, plosive, labial, dental and alveolar features - have close to zero weights. Similarly to the N\&M dataset, the palatal feature is shared by: /\textipa{S}/, /\textipa{tS}/, and /\textipa{dZ}/, which are all distributed-strident consonants, and also by /\textipa{j}/. As for the glide, lateral, and rhotic features, the corresponding consonant phonemes are /\textipa{w}/ and /\textipa{j}/ (glide phonemes) and /\textipa{l}/ and /\textipa{r}/ (liquid phonemes). These four phonemes are characterized as approximant consonants, as the articulators are closer to each other than for vowels, without creating turbulence in the airflow as is the case with fricative consonants. 

\paragraph{Phonological features} For the phonological features, the resulting order of the subphonemic features is: voicing, nasality, strident, consonantal, lateral, coronal and continuant. All differences are statistically significant (t-test $p$-value$<0.05$) except for the lateral-coronal difference. The other features - dorsal, anterior and labial - have weights close to zero. This order follows a similar pattern to the one described so far: voicing, nasality, distributed-stridents and approximants. The two leading features are again voicing and nasality. The stridency feature, and the consonantal and lateral features, which contrast between the approximant phonemes /j/, /w/ and /l/ to other phonemes, all have relatively high weights also in this case. We therefore summarize the order between subphonemic features, based on all the results so far, as: voicing, nasality, distributed-stridents and approximants. As for the coronal and continuant features, note that they are partially redundant with respect to the leading ones and do not clearly contrast among phoneme sets. For example, the coronal feature is shared by both strident phonemes (/\textipa{s}/, /\textipa{z}/, /\textipa{S}/, /\textipa{tS}/, /\textipa{dZ}/), approximants (/\textipa{l}/, /\textipa{r}/, /\textipa{j}/) and nasal (/\textipa{n}/) phonemes, but also by /\textipa{t}/, /\textipa{d}/, /\textipa{T}/, /\textipa{D}/.

\paragraph{A two-dimensional visualization of perceived distances} To visualize the high-dimensional structure of the perceived distances in the datasets, and the relations between subphonemic features, we use Multi-Dimensional Scaling (MDS) \citep{kruskal1964}, which embeds all perceived distances on the plane. This allows us to \textit{qualitatively} assess if the learned feature weights are indeed reflected in the visualization of the perceptual distances.

\begin{figure}[ht]
\makebox[\linewidth][c]{\includegraphics[width=10cm]{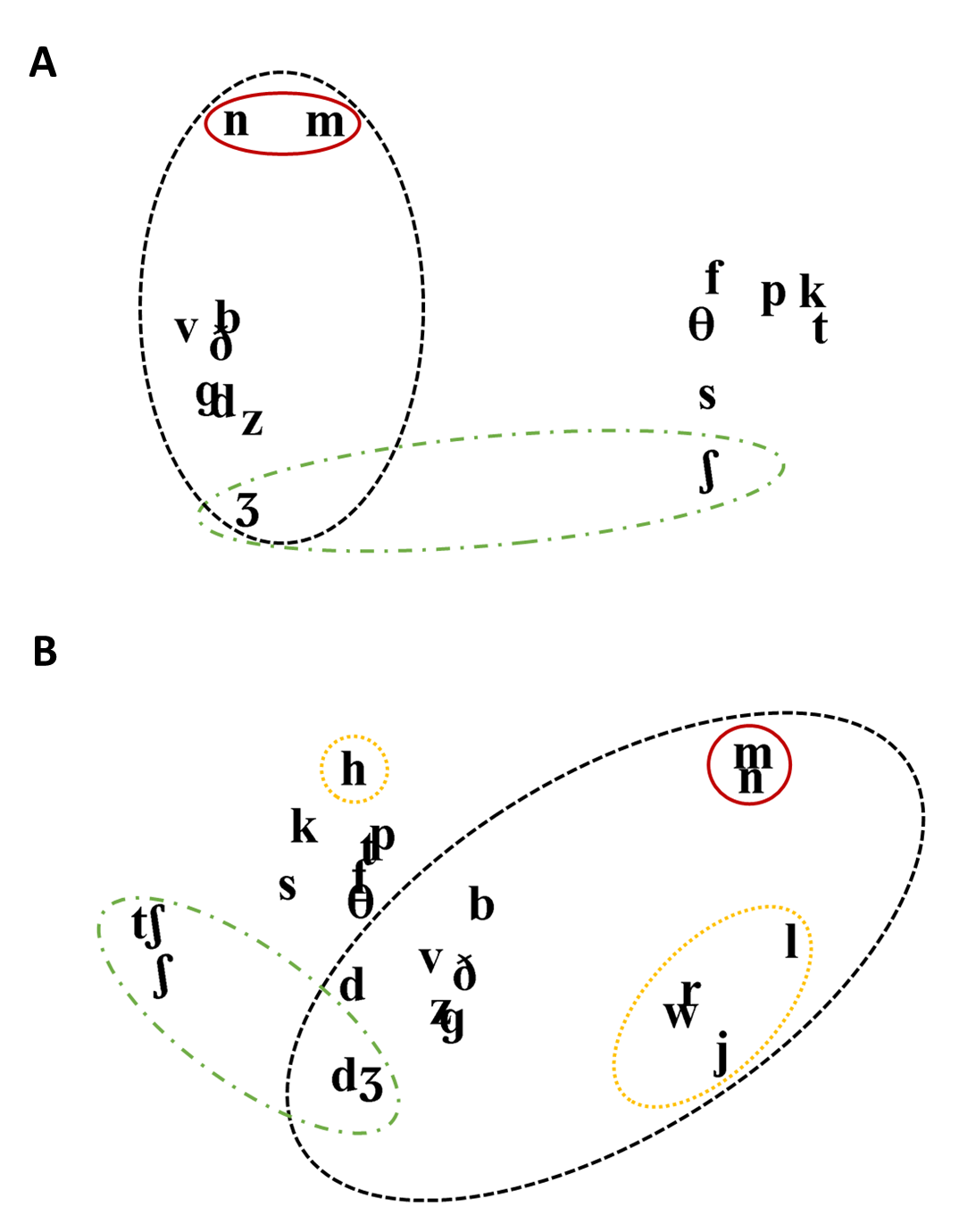}}
\caption{Multi-Dimensional Scaling of perceptual distances between phonemes for the  (A) N\&M datset (B) Luce dataset. Ovals: black (dashed line) - voiced phonemes, red (solid line) - nasals, green - distributed stridents (dash-dot line), yellow (dotted line)- approximants.}.
\end{figure}

MDS is a method for embedding objects and their pairwise distances onto a lower-dimensional space, where the distances between the objects in the lower-dimensional plane approximate the original distances. MDS was already used to explore the N\&M and other phoneme-confusion dataset \citep{Shepard1980, mielke2008emergence, Buchsbaum2015}, we test it also for the Luce dataset. Figure 5 presents the results of embedding the perceptual distances between all phonemes into a 2D-plane for the N\&M dataset (panel A) and for the Luce dataset (panel B). The classes of the resulting four leading features are marked with colored ovals: voiced phonemes with black (dashed line), nasal phonemes with red (solid line), distributed-strident phonemes with green (dash-dot line), and approximant-consonants with yellow (dotted line) - the last is only for the Luce dataset, as there were no approximant phonemes in the N\&M experiment. 

The MDS analysis provides several insights. First, for both datasets, voiced and voiceless phonemes reside at opposite sides of the plane. Second, nasal, distributed-strident and approximant phonemes are in the periphery of the plot, and are relatively distant from other phonemes. A phoneme will tend to result in the periphery when it is relatively distant from all other phonemes, which should match a high weight of the corresponding features. Finally, a clear distinction between distributed and non-distributed stridents is observed for both datasets. Among all stridents only the distributed stridents - (/\textipa{S}/, /\textipa{Z}/) in the N\&M dataset, and (/\textipa{tS}/, /\textipa{dZ}/ and /\textipa{S}/) in the Luce dataset - are in the periphery of the plot, whereas the non-distributed stridents (/\textit{s}/, /\textipa{z}/) are more central. This corresponds to the relatively high weights of the palatal feature that is shared by the distributed stridents but not by the non-distributed stridents.

\paragraph{Redundancy between features} To further examine the resulting feature weights, we tested how omitting a feature from the feature theory would influence the predictive power of the model and the resulting order between subphonemic features. We hypothesized that leaving out features that have relatively high weights would reduce the predictive power of the model and would affect the resulting order.

We therefore repeated the above analyses, leaving out a different feature from the representation scheme on each step, and finally estimating the new prediction of the reduced model. We present results for the larger dataset, the Luce dataset: Figure 6 presents the differences between the prediction of the full model and the reduced model, in ascending order of size, for each of the left-out features from the representation scheme. The prediction difference is shown for the two feature theories: articulatory features (panel A) and phonological features (panel B). The prediction difference is significant between the voicing and nasality features compared to all other features (t-test $p$-value$<0.05$), whereas results for the voicing and nasality features are statistically comparable. The palatal feature is statistically comparable to the approximant features - glide, lateral and rhotic features - and to the glottal and fricative features, and significantly different compared to the rest of the features.

\begin{figure}[ht]
\vspace{.3in}
\makebox[\textwidth][c]{\includegraphics[width=\textwidth]{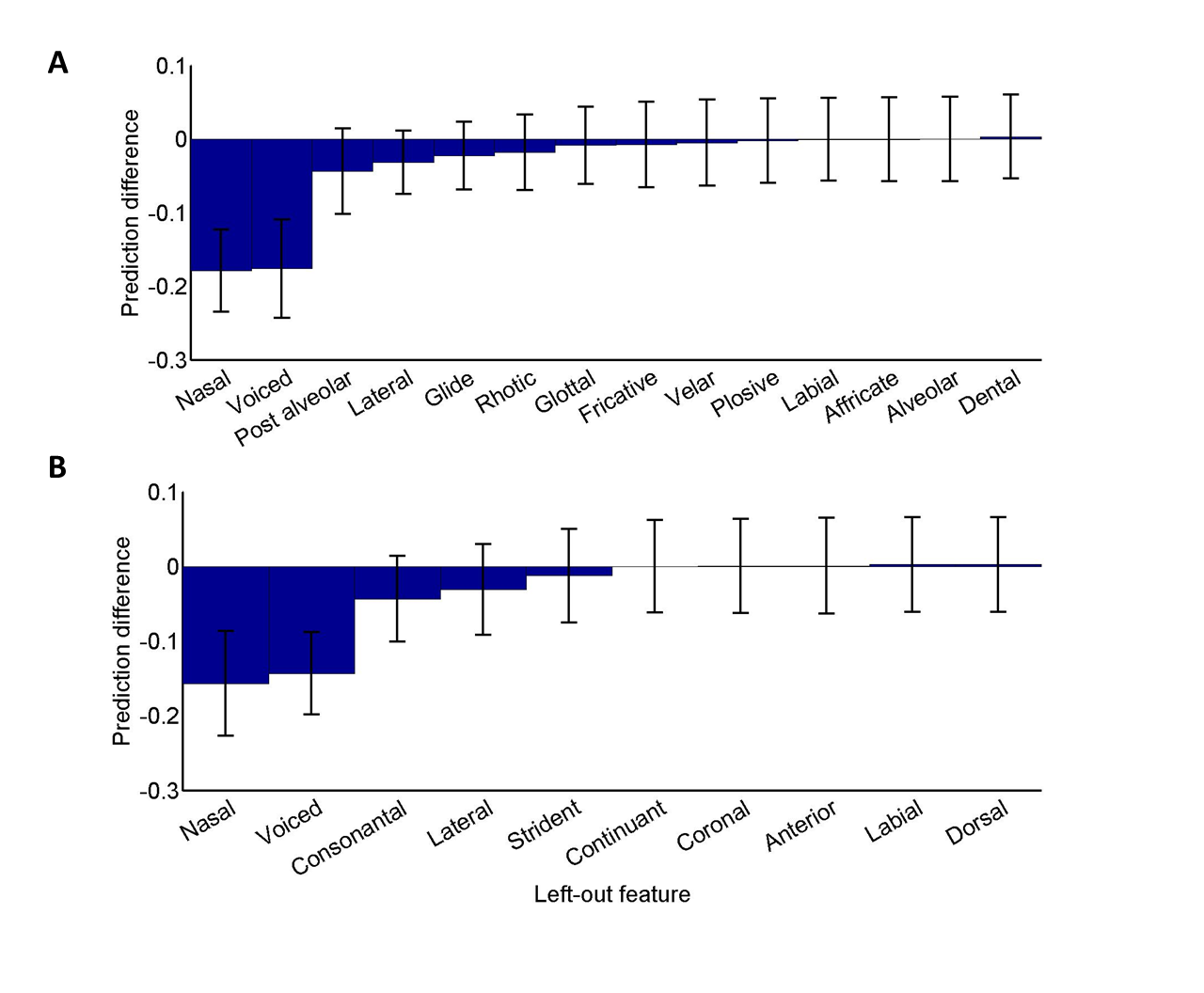}}
\caption{Reduction in model prediction as a result of omitting a single feature, as calculated for the Luce dataset. (A) Articulatory features (B) Phonological features.}
\end{figure}

Results show that omitting high-weight features significantly reduces the predictive power of the model. In contrast, omitting low-weight features hardly affects prediction.

\subsection{Phoneme confusion in Hebrew and English}
This section analyzes the confusion data of Hebrew in the same way as for English in section 4.2. Hebrew phonemes are represented based on two feature theories as described above, and the LS-diagonal method is applied for each of these cases.

\begin{figure}[ht]
\vspace{.3in}
\includegraphics[width=\linewidth]{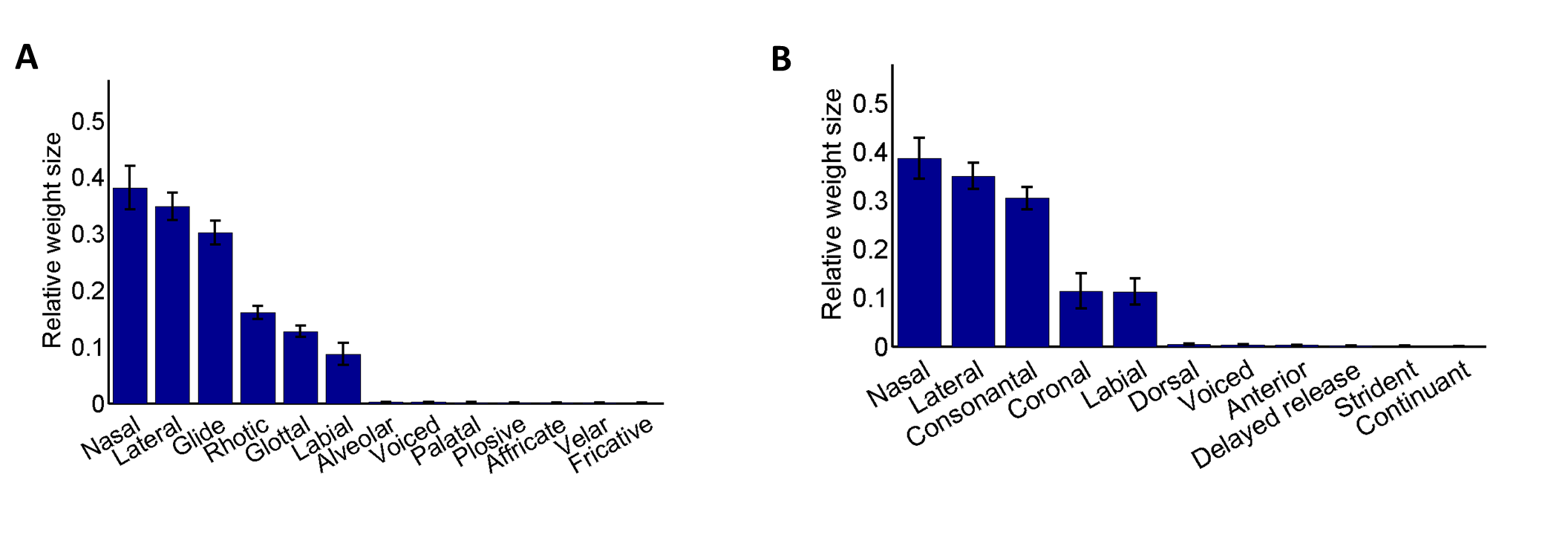}
\caption{Weights of subphonemic features as derived from the Hebrew dataset. (A) Articulatory features (B) Phonological features. Error bars represent SD across validation sets.}
\end{figure}

Figure 7 shows the resulting feature weights for articulatoy and phonological features. Computed for the Hebrew dataset, results show that as in English datasets, the nasality feature and the approximant features - lateral, glide and rhotic - are heavily weighted. The voicing feature, on the other hand, has much lower weight in Hebrew compared to English. In addition, the labial feature in both articulatory and phonological feature theories has relatively high weight compared to the English results.

The results of MDS analysis on the Hebrew dataset (Figure 8A) agree with the results of the metric learning. Nasal phonemes are embedded in the periphery of the plot, as well as the approximants and the glottal phoneme /h/. Unlike the N\&M and Luce embeddings, voiced phonemes are not well separated from unvoiced phonemes. The locations of the labial phonemes /p/ and /f/ in the MDS analysis is peripheral, which corresponds to the relatively high weight of the labial phoneme. 

\begin{figure}[hb]
\vspace{.3in}
\includegraphics[width=\linewidth]{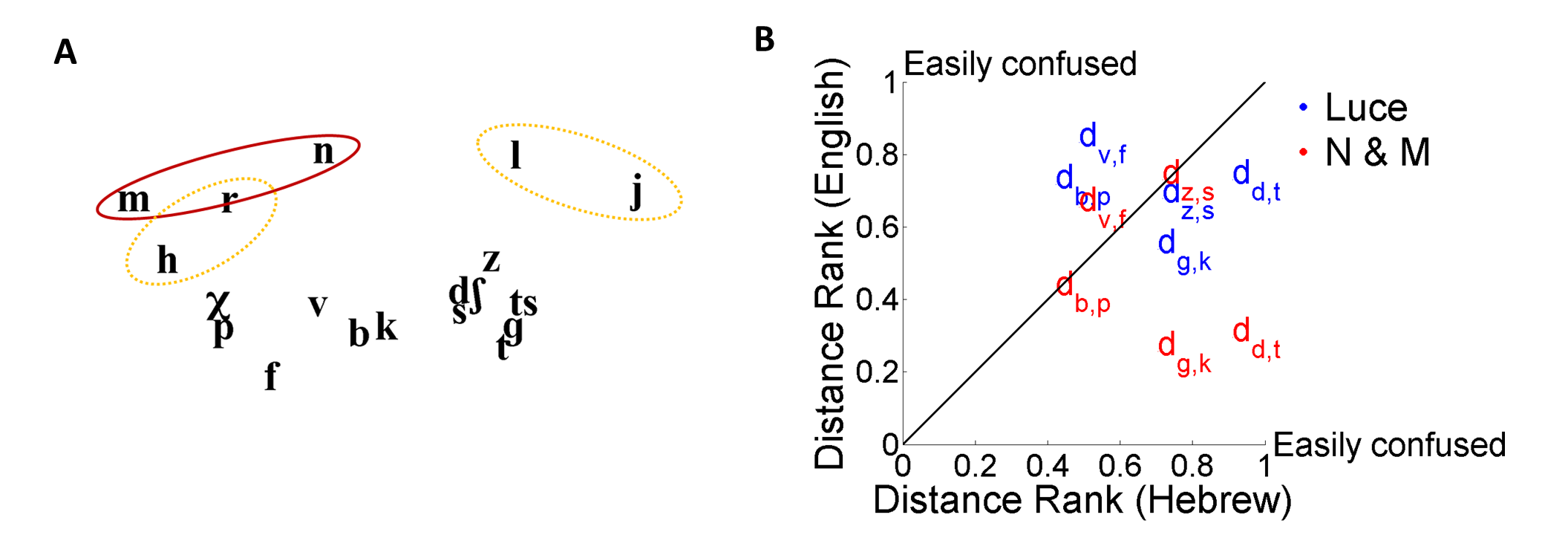}
\caption{(A) Multi-Dimensional Scaling for the Hebrew dataset. Ovals: red (solid line) - nasals, yellow (dotted line)- approximants and the glottal /h/(B) Voicing-differing minimal pairs: Hebrew vs. English}
\end{figure}

Figure 8B summarizes the differences between the Luce dataset and Hebrew with respect to perceptually-discriminative power of subphonemic features. Values on each axis represent normalized weights. Subphonmeic features that are below the line $y = x$ are more perceptually discriminative in Hebrew, and vice versa. Note that the farthest feature from the line is the voicing feature. Given this large discrepancy with respect to voicing between English and Hebrew, we further analyze this difference in the following subsection.

\subsubsection{Voicing-differing minimal pairs} To further examine the difference between Hebrew and English with respect to voicing, we compare a set of minimal pairs that only differ in voicing. We extract perceived pairwise distances between (/b/-/p/, /d/-/t/, /g/-/k/, /z/-/s/ and /v/-/f/) from the three datasets, and compare their rank on a scatter plot (figure 9). Ranking is necessary to equalize the distributions across languages: the values on the axes represent the rank of the perceived distance of a pair among all other pairwise distances. Before ranking distances, the largest subset of phonemes shared by all three datasets (13 phonemes) was first extracted. This was done to ensure that the rank of the distance does not depend on the specific phoneme inventory of the dataset. Features that corresponds to points above the line $y=x$ are more confusable in Hebrew than in English, and vice versa. The distance of a point from this line can serve as a measure of the difference between the two languages, with respect to the voicing difference between the phonemes in the minimal pair.

There are ten pairwise distances, five for the contrast between Hebrew and the Luce dataset (blue), and five for the contrast between Hebrew and the N\&M dataset (red). Among the ten points, five points are below the dividing line, three points above, and two reside on the line. There is a bias towards more confusion in Hebrew compared to English, with respect to voiced/voiceless phoneme confusion. This agrees with the relatively low weight of voicing in Hebrew. However, looking at the distribution of distances from the dividing line, this bias is not statistically significant (Wilcoxon rank test $p$-value$=0.57$). A possible confound for this is the relatively high weight of the labial feature in Hebrew, in particular, the relatively large distance of /p/ and /f/ from other phonemes as can be observed in the MDS analysis. This could counteract the difference between English and Hebrew with respect to voicing. Looking at the distribution of the non-labial phonemes (/d/-/t/, /g/-/k/ and /s/-/z/) the bias towards Hebrew is significant (Wilcoxon rank test $p$-value$<0.05$).

\begin{figure}[h]
\vspace{.3in}
\includegraphics[width=\linewidth]{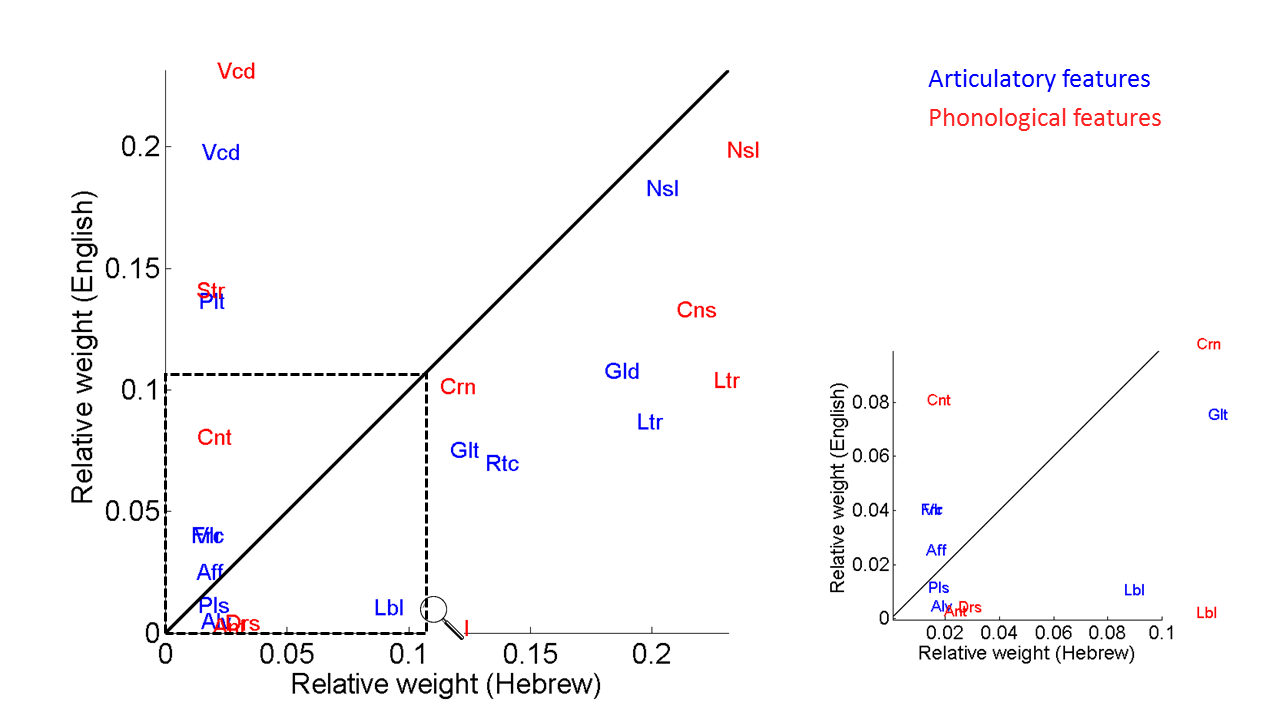}
\caption{English vs. Hebrew: relative weights of features. Articulatory features in blue, Phonological features in red. Dashed square indicates the magnified area (bottom-right). Features that corresponds to points above the line $y=x$ are more confusable in Hebrew than in English, and vice versa. Feature names are abbreviated according to their three first consonants in their name.}
\end{figure}
\section{Summary and discussion}
This study explored a new framework to the problem of phoneme similarity. Using learning algorithms, the suggested framework derives a feature-based metric function from phoneme-confusion data. In doing so, it provides means to quantify the contribution of various subphonemic features to the overall similarity. Moreover, since feature weights compete among themselves over explaining empirical perceptual distances (section 2.1), the resulting weights are interpreted as perceptual saliencies of features.

We explored with two classic phoneme-confusion datasets in English \citep{NicelyMiller1955, Luce1987} and two common feature theories (\citealp{ChomskyHalle1968}, and articulatory dimensions). We then derived a metric function for each dataset and feature theory, and described the resulting order among subphonemic features with respect to their perceptual saliencies. The four leading groups of subphonemic features in English, in descending order, are: voicing, nasality, distributed-stridents, and approximants. Desirably, this is in accordance with previous studies and acoustic considerations: the voicing and nasality features have been demonstrated as highly perceptually discriminating using information-theory measures \citep{NicelyMiller1955}. As for distributed-stridents, this class of phonemes have relatively high energy in their waveform, which may explain their relatively high weight. Acoustic analyses conducted by \citep[see][Figure 15]{Mielke2012} showed that stridents are distinguished from other phonemes along the first principal component of acoustic distances, which is interpreted as dominance of high-frequency acoustic energy. However, a distinction between distributed stridents and non-distributed stridents was not observed in this analysis. The second principal component of this analysis, which is interpreted as dominance of low-frequency acoustic energy, distinguishes the nasal and approximant phonemes /n/, /m/, /w/, /j/, /l/, /r/ and /h/ from the other phonemes in the datasets, in this order. This shows that the relatively high weight of approximant features and the fricative glottal /h/ is also grounded in acoustics. Taken together, the results of this study for the relative perceptual saliency of phonological features concur with previous studies, and extend them in several aspects. First, the results provide a more detailed view on the order among phonological features with respect to their perceptual saliency, with four leading classes for English: voicing, nasality, distributed-stridents and approximants. Second, the suggested framework provides a \textit{quantification} of perceptual saliency, in contrast to qualitative differences. This can be beneficial for subsequent studies, on, e.g., speech perception or verbal memory, which require means to quantify perceptual saliencies of individual phonological features. Finally, in contrast to previous studies on perceptual similarity \citep{Tversky1977, Pierrehumbert1993, Frisch1997}, the suggested framework derives the similarity function in a data-driven manner, rather than manually, showing a better generalization to unseen data.

In addition, we report on confusion-rate results among Hebrew phonemes. The new Hebrew dataset was analyzed and compared to the English ones, revealing a discrepancy between Hebrew and English with respect to subphonemic feature weights. In particular, a major difference with respect to voicing was found. We suggest two possible sources for this difference. First, it can be accounted for by redundant cues for unvoiced plosives in English that are absent in Hebrew. A redundant cue is an articulatory-independent mechanism to enhance a primary cue in the signal. In particular, \citet{stevens1989primary} suggests that the feature [-voice] is enhanced by glottal opening (aspiration) in English, which prolongs the duration of the voiceless interval, thus rendering voiceless stops more distinguishable from voiced stops. Hebrew and other languages (e.g., French) do not employ this enhancement. Another explanation for the different weight of voicing is the differences in voicing onset times (VOT) between the two languages \citep{Laufer1998}. VOTs of voiced stops in Hebrew are more negative than in English, with mean values close to 100ms compared to 60ms in English. Such differences may affect perceptual discrimination of the voicing feature. Taken together, this result is another evidence that the metric space may differ across languages and emphasizes the need for deriving a separate metric function for each language, as our framework provides.

The study relies on several simplifying assumptions that can be investigated in future studies: for computational convenience, it assumes a linear mapping between the feature and similarity spaces. However, the suggested framework provides freedom in the choice of the general form of the metric function, which can incorporate any type of non-linearity (Eq. 1). Second, the study assumes that perceptual similarities are symmetric, despite accumulating opposing evidence. However, asymmetric metrics, namely - quasimetrics, can be learned using the same approach. Once the form of the quasimetric is chosen, its parameters can be learned from the data, in the same way, subject to minor modifications of the learning algorithms. 

In sum, this study is a first demonstration of the benefits of a data-driven approach to the problem of phoneme similarity. Its general framework was shown to have several potential advantages, and provides means to quantify various aspects of phoneme similarity. In particular, it reveals an order among subphonemic features with respect to their perceptually discriminative power, suggesting the first quantification of their contribution to the overall similarity. The approach enables the integration of various effects, such as context or asymmetry, into its framework, and is compatible with various feature theories and noise characteristics.

\subsubsection*{Acknowledgements}
We thank Aviad Albert for his essential help and comments. Computing equipment for this work was supported by the Israeli Science Foundation for GC (1090/2012),  and by an NVidia research grant to GC.

\bibliography{bib_thesis.bib}  
\end{document}